\documentclass{article}
\usepackage{hltnaacl03}

\title{Introduction to the CoNLL-2003 Shared Task:
       Language-Independent Named Entity Recognition}
\author{
{\bf Erik F. Tjong Kim Sang} \and {\bf Fien De Meulder}\\
CNTS - Language Technology Group\\
University of Antwerp\\
{\tt \{erikt,fien.demeulder\}@uia.ua.ac.be}
}

\date{\today}

\begin{document}
\maketitle

\begin{abstract}
We describe the CoNLL-2003 shared task: language-independent named
entity recognition.
We give background information on the data sets (English and German)
and the evaluation method, present a general overview of the systems
that have taken part in the task and discuss their performance.
\end{abstract}

\section{Introduction}

Named entities are phrases that contain the names of persons,
organizations and locations. Example:

\begin{quote}
[ORG U.N. ] official [PER Ekeus ] heads for [LOC Baghdad ] .
\end{quote}

\noindent
This sentence contains three named entities: 
{\it Ekeus} is a person,
{\it U.N.} is a organization and
{\it Baghdad} is a location.
Named entity recognition is an important task of information
extraction systems.
There has been a lot of work on named entity recognition, especially 
for English (see Borthwick \shortcite{borthwick99} for an overview). 
The Message Understanding Conferences (MUC) have offered developers 
the opportunity to evaluate systems for English on the same data in 
a competition.
They have also produced a scheme for entity annotation
\cite{chinchor99}.
More recently, there have been other system development competitions
which dealt with different languages 
(IREX and CoNLL-2002).

The shared task of CoNLL-2003 concerns language-independent named
entity recognition. 
We will concentrate on four types of named entities: persons,
locations, organizations and names of miscellaneous entities that do
not belong to the previous three groups. 
The shared task of CoNLL-2002 dealt with named entity recognition for
Spanish and Dutch \cite{tjongintro2002}.
The participants of the 2003 shared task have been offered training
and test data for two other European languages: English and German.
They have used the data for developing a named-entity recognition
system that includes a machine learning component.
The shared task organizers were especially interested in approaches
that made use of resources other than the supplied training data, for
example gazetteers and unannotated data.

\section{Data and Evaluation}

In this section we discuss the sources of the data that were used
in this shared task, the preprocessing steps we have performed on the
data, the format of the data and the method that was used for
evaluating the participating systems.

\subsection{Data}

The CoNLL-2003 named entity data consists of eight files covering two
languages: English and 
German\footnote{
Data files (except the words) 
can be found
on http://lcg-www.uia.ac.be/conll2003/ner/}.
For each of the languages there is a training file, a development
file, a test file and a large file with unannotated data.
The learning methods were trained with the training data.
The development data could be used for tuning the parameters of the
learning methods.
The challenge of this year's shared task was to incorporate the
unannotated data in the learning process in one way or another.
When the best parameters were found, the method could be trained on the
training data and tested on the test data.
The results of the different learning methods on the test sets are 
compared in the evaluation of the shared task.
The split between development data and test data was chosen to
avoid systems being tuned to the test data.

The English data was taken from the Reuters Corpus\footnote{
  http://www.reuters.com/researchandstandards/}. This corpus
consists of Reuters news stories between August 1996 and August 1997.
For the
training and development set, ten days' worth of data were taken from
the files representing the end of August 1996. For the test set, the
texts were from December 1996. The preprocessed raw data
covers the month of September 1996.

The text for the German data was taken from the ECI Multilingual Text
Corpus\footnote{
http://www.ldc.upenn.edu/}.
This corpus consists of texts in many languages.
The portion of data that was used for this task, was extracted from
the German newspaper Frankfurter Rundshau. All three of the training,
development and test sets were taken from articles written in one week
at the end of August 1992. The raw data were taken from the months of
September to December 1992.

Table~\ref{corpusstats} contains an overview of the sizes of the data
files.
The unannotated data contain 17 million tokens (English) and
14 million tokens (German).


\begin{table}[t]
\begin{center}
\begin{tabular}{|l|c|c|c|}\cline{2-4}
\multicolumn{1}{l|}{English data} & Articles & Sentences & Tokens \\\hline
Training set    & 946 & 14,987 & 203,621 \\           
Development set & 216 &  3,466 & 51,362 \\ 
Test set        & 231 &  3,684 & 46,435 \\\hline
\multicolumn{3}{c}{}\\\cline{2-4}
\multicolumn{1}{l|}{German data} & Articles & Sentences & Tokens \\\hline
Training set    & 553 & 12,705 & 206,931 \\           
Development set & 201 &  3,068 & 51,444 \\ 
Test set        & 155 &  3,160 & 51,943 \\\hline
\end{tabular}
\end{center}
\caption{
Number of articles, sentences and tokens in each data file.
} 
\label{corpusstats}
\end{table}

\subsection{Data preprocessing}

The participants were given access to the corpus after some linguistic
preprocessing had been done: for all data, a tokenizer, part-of-speech
tagger, and a chunker were applied to the raw data. We created two 
basic language-specific tokenizers for this shared task.
The English data was tagged and chunked by the memory-based MBT
tagger \cite{mbt2002}.
The German data was lemmatized, tagged and chunked by the decision
tree tagger Treetagger \cite{schmid95}.

Named entity tagging of English and German training, development, and
test data, was done by hand at the University of Antwerp. Mostly, MUC
conventions were followed \cite{chinchor99}. An extra named entity
category called MISC was added to denote all names which are not
already in the other categories. This includes adjectives, like {\it
Italian}, and events, like {\it 1000 Lakes Rally}, making it a very
diverse category.

\begin{table}[t]
\begin{center}
\begin{tabular}{|l|c|c|c|c|}\cline{2-5}
\multicolumn{1}{l|}{English data} & LOC & MISC & ORG & PER \\\hline
Training set    & 7140 & 3438 & 6321 & 6600 \\
Development set & 1837 &  922 & 1341 & 1842 \\
Test set        & 1668 &  702 & 1661 & 1617 \\\hline
\multicolumn{3}{c}{}\\\cline{2-5}
\multicolumn{1}{l|}{German data} & LOC & MISC & ORG & PER \\\hline
Training set    & 4363 & 2288 & 2427 & 2773 \\
Development set & 1181 & 1010 & 1241 & 1401 \\
Test set        & 1035 &  670 &  773 & 1195 \\\hline
\end{tabular}
\end{center}
\caption{
Number of named entities per data file
} 
\label{tab-nestats}
\end{table}

\subsection{Data format}

All data files contain one word per line
with empty lines representing sentence boundaries.
At the end of each line there is a tag which states whether the 
current word is inside a named entity or not.
The tag also encodes the type of named entity.
Here is an example sentence:

\begin{quote}
\begin{tabular}{rlll}
U.N.        & NNP & I-NP & I-ORG \\
official    & NN  & I-NP & O \\
Ekeus       & NNP & I-NP & I-PER \\
heads       & VBZ & I-VP & O \\
for         & IN  & I-PP & O \\
Baghdad     & NNP & I-NP & I-LOC \\
.           & .   & O    & O \\
\end{tabular}
\end{quote}

\noindent
Each line contains four fields: the word, its part-of-speech tag, its
chunk tag and its named entity tag.
Words tagged with O are outside of named entities and
the I-XXX tag is used for words inside a named entity of type XXX.
Whenever two entities of type XXX are immediately next to each
other, the first word of the second entity will be tagged B-XXX
in order to show that it starts another entity.
The data contains entities of four types:
persons (PER),
organizations (ORG),
locations (LOC) and
miscellaneous names (MISC).
This tagging scheme is the IOB scheme originally put forward by
Ramshaw and Marcus \shortcite{ramshaw95}. 
We assume that named entities are non-recursive and non-overlapping.
When a named entity is embedded in another named entity, usually
only the top level entity has been annotated.

Table~\ref{tab-nestats} contains an overview of the number of named
entities in each data file.

\begin{table*}[t]
\begin{center}
\begin{tabular}{|l|c|c|c|c|c|c|c|c|c|c|c|c|c|c|}\cline{2-14}
\multicolumn{1}{l|}{}
             & lex & pos & aff & pre & ort & gaz & chu & pat & cas & tri & bag & quo & doc \\\hline
Florian      & +   & +   & +   & +   & +   & +   & +   & -   & +   & -   & -   & -   & -   \\
Chieu        & +   & +   & +   & +   & +   & +   & -   & -   & -   & +   & -   & +   & +   \\
Klein        & +   & +   & +   & +   & -   & -   & -   & -   & -   & -   & -   & -   & -   \\
Zhang        & +   & +   & +   & +   & +   & +   & +   & -   & -   & +   & -   & -   & -   \\
Carreras (a) & +   & +   & +   & +   & +   & +   & +   & +   & -   & +   & +   & -   & -   \\
Curran       & +   & +   & +   & +   & +   & +   & -   & +   & +   & -   & -   & -   & -   \\
Mayfield     & +   & +   & +   & +   & +   & -   & +   & +   & -   & -   & -   & +   & -   \\
Carreras (b) & +   & +   & +   & +   & +   & -   & -   & +   & -   & -   & -   & -   & -   \\
McCallum     & +   & -   & -   & -   & +   & +   & -   & +   & -   & -   & -   & -   & -   \\
Bender       & +   & +   & -   & +   & +   & +   & +   & -   & -   & -   & -   & -   & -   \\
Munro        & +   & +   & +   & -   & -   & -   & +   & -   & +   & +   & +   & -   & -   \\
Wu           & +   & +   & +   & +   & +   & +   & -   & -   & -   & -   & -   & -   & -   \\
Whitelaw     & -   & -   & +   & +   & -   & -   & -   & -   & +   & -   & -   & -   & -   \\
Hendrickx    & +   & +   & +   & +   & +   & +   & +   & -   & -   & -   & -   & -   & -   \\
De Meulder   & +   & +   & +   & -   & +   & +   & +   & -   & +   & -   & -   & -   & -   \\
Hammerton    & +   & +   & -   & -   & -   & +   & +   & -   & -   & -   & -   & -   & -   \\\hline
\end{tabular}
\end{center}
\caption{
Main features used by the the sixteen systems
that participated in the CoNLL-2003 shared task
sorted by performance on the English test data.
Aff: affix information (n-grams);
bag: bag of words;
cas: global case information;
chu: chunk tags;
doc: global document information;
gaz: gazetteers;
lex: lexical features;
ort: orthographic information;
pat: orthographic patterns (like Aa0);
pos: part-of-speech tags;
pre: previously predicted NE tags;
quo: flag signing that the word is between quotes;
tri: trigger words.
}
\label{tab-feat}
\end{table*}

\subsection{Evaluation}

The performance in this task is measured with F$_{\beta=1}$ rate:

\begin{equation}
F_{\beta} = \frac{(\beta^2+1)*precision*recall}
                 {(\beta^2*precision+recall)}
\end{equation}

\noindent
with $\beta$=1 \cite{vanrijsbergen75}.
Precision is the percentage of named entities found by the learning
system that are correct.
Recall is the percentage of named entities present in the corpus that
are found by the system.
A named entity is correct only if it is an exact match of the
corresponding entity in the data file.

\section{Participating Systems}

Sixteen systems have participated in the CoNLL-2003 shared task.
They employed a wide variety of machine learning techniques as well as
system combination.
Most of the participants have attempted to use information other than
the available training data.
This information included gazetteers and unannotated data, and there
was one participant who used the output of externally trained
named entity recognition systems.

\subsection{Learning techniques}

The most frequently applied technique in the CoNLL-2003 shared task is
the Maximum Entropy Model.
Five systems used this statistical learning method.
Three systems used Maximum Entropy Models in isolation
\cite{bender2003,chieu2003,curran2003}.
Two more systems used them in combination with other techniques
\cite{florian2003,klein2003}.
Maximum Entropy Models seem to be a good choice for this kind of task:
the top three results for English and the top two results for German
were obtained by participants who employed them in one way or another.

Hidden Markov Models were employed by four of the systems that took
part in the shared task
\cite{florian2003,klein2003,mayfield2003,whitelaw2003}.
However, they were always used in combination with other learning
techniques. 
Klein et al. \shortcite{klein2003} also applied the related 
Conditional Markov Models for combining classifiers.

Learning methods that were based on connectionist approaches were
applied by four systems.
Zhang and Johnson \shortcite{zhang2003} used robust risk minimization, 
which is a Winnow technique.
Florian et al. \shortcite{florian2003} employed the same technique in
a combination of learners.
Voted perceptrons were applied to the shared task data by
Carreras et al. \shortcite{carreras2003b} and Hammerton used a recurrent
neural network (Long Short-Term Memory) for finding named entities.

Other learning approaches were employed less frequently.
Two teams used AdaBoost.MH \cite{carreras2003a,wu2003} and two other
groups employed memory-based learning
\cite{demeulder2003,hendrickx2003}. 
Transformation-based learning \cite{florian2003},
Support Vector Machines \cite{mayfield2003} and
Conditional Random Fields \cite{mccallum2003} were applied by one
system each.

Combination of different learning systems has proven to be a good
method for obtaining excellent results.
Five participating groups have applied system combination.
Florian et al. \shortcite{florian2003} tested different 
methods for combining the results of four systems and found that
robust risk minimization worked best.
Klein et al. \shortcite{klein2003} employed a stacked learning system
which contains Hidden Markov Models, Maximum Entropy Models and
Conditional Markov Models. 
Mayfield et al. \shortcite{mayfield2003} stacked two learners and
obtained better performance.
Wu et al. \shortcite{wu2003} applied both stacking and voting to three
learners. 
Munro et al. \shortcite{munro2003} employed both voting and bagging
for combining classifiers. 

\subsection{Features}

The choice of the learning approach is important for obtaining a
good system for recognizing named entities.
However, in the CoNLL-2002 shared task we found out that choice of
features is at least as important.
An overview of some of the types of features
chosen by the shared task participants, can be found in
Table~\ref{tab-feat}.

All participants used lexical features (words) except for Whitelaw and
Patrick \shortcite{whitelaw2003} who implemented a character-based
method. 
Most of the systems employed part-of-speech tags and two of them have
recomputed the English tags with better taggers
\cite{hendrickx2003,wu2003}.
Othographic information, affixes, gazetteers and chunk information
were also incorporated in most systems although one group reports that
the available chunking information did not help
\cite{wu2003}
Other features were used less frequently.
Table~\ref{tab-feat} does not reveal a single feature that would be
ideal for named entity recognition.

\subsection{External resources}

Eleven of the sixteen participating teams have attempted
to use information other than the training data that was
supplied for this shared task.
All included gazetteers in their systems.
Four groups examined the usability of unannotated data,
either for extracting training instances
\cite{bender2003,hendrickx2003} or obtaining extra named
entities for gazetteers \cite{demeulder2003,mccallum2003}.
A reasonable number of groups have also employed unannotated data for
obtaining capitalization features for words.
One participating team has used externally trained named entity
recognition systems for English as a part in a combined system
\cite{florian2003}.

Table~\ref{tab-extrainfo} shows the error reduction of the systems
with extra information compared to while using only the available
training data.
The inclusion of extra named entity recognition systems seems to have
worked well \cite{florian2003}.
Generally the systems that only used gazetteers seem to gain more than
systems that have used unannotated data for other purposes than
obtaining capitalization information.
However, the gain differences between the two approaches are most
obvious for English, for which better gazetteers are available.
With the exception of the result of Zhang and Johnson
\shortcite{zhang2003}, there is not much difference in the German
results between the gains obtained by using gazetteers and those
obtained by using unannotated data. 

\begin{table}[t]
\begin{center}
\begin{tabular}{|l|c|c|c|c|c|}\cline{2-6}
\multicolumn{1}{l|}{}
                     & G & U & E & English & German \\\hline
Zhang                & +   & -   & -   & 19\% & 15\% \\
Florian              & +   & -   & +   & 27\% &  5\% \\
Hammerton            & +   & -   & -   & 22\% &    - \\
Carreras        (a)  & +   & -   & -   & 12\% &  8\% \\
Chieu                & +   & -   & -   & 17\% &    - \\
Hendrickx            & +   & +   & -   &  7\% &  5\% \\
De Meulder           & +   & +   & -   &  8\% &  3\% \\
Bender               & +   & +   & -   &  3\% &  6\% \\
Curran               & +   & -   & -   &  1\% &    - \\
McCallum             & +   & +   & -   &    ? &    ? \\
Wu                   & +   & -   & -   &    ? &    ? \\\hline
\end{tabular}
\end{center}
\caption{
Error reduction for the two development data sets when using extra
information like gazetteers (G), unannotated data (U) or externally
developed named entity recognizers (E).
The lines have been sorted by the sum of the reduction percentages for
the two languages.
}
\label{tab-extrainfo}
\end{table}

\subsection{Performances}

A baseline rate was computed for the English and the German test
sets.
It was produced by a system which only identified entities which had a
unique class in the training data.
If a phrase was part of more than one entity, the system would select
the longest one.
All systems that participated in the shared task have outperformed
the baseline system.

For all the F$_{\beta=1}$ rates we have estimated significance
boundaries by using bootstrap resampling \cite{noreen89}.
From each output file of a system, 250 random samples of sentences
have been chosen and the distribution of the F$_{\beta=1}$ rates in
these samples is assumed to be the distribution of the performance
of the system.
We assume that performance A is significantly different from
performance B if A is not within the center 90\% of the distribution
of B.

The performances of the sixteen systems on the two test data sets can
be found in Table~\ref{tab-results}.
For English, the combined classifier of Florian et al.
\shortcite{florian2003} achieved the highest overall F$_{\beta=1}$
rate.  
However, the difference between their performance and that of the
Maximum Entropy approach of Chieu and Ng \shortcite{chieu2003} 
is not significant.
An important feature of the best system that other participants
did not use, was the inclusion of the output of two externally trained
named entity recognizers in the combination process.
Florian et al. \shortcite{florian2003} have also obtained the highest
F$_{\beta=1}$ rate for the German data.
Here there is no significant difference between them and the systems
of Klein et al. \shortcite{klein2003} and Zhang and Johnson
\shortcite{zhang2003}.

We have combined the results of the sixteen system in order to see if
there was room for improvement.
We converted the output of the systems to the same IOB tagging
representation and searched for the set
of systems from which the best tags for the development data could be
obtained with majority voting.
The optimal set of systems was determined by performing a
bidirectional hill-climbing search \cite{caruana94} with beam size 9,
starting from zero features.
A majority vote of five systems
\cite{chieu2003,florian2003,klein2003,mccallum2003,whitelaw2003}
performed best on the English development data.
Another combination of five systems
\cite{carreras2003a,mayfield2003,mccallum2003,munro2003,zhang2003}
obtained the best result for the German development data.
We have performed a majority vote with these sets of systems on the
related test sets and obtained F$_{\beta=1}$ rates of
90.30 for English (14\% error reduction compared with the best system)
and
74.17 for German (6\% error reduction).


\section{Concluding Remarks}

We have described the CoNLL-2003 shared task: language-independent
named entity recognition.
Sixteen systems have processed English and German named entity data.
The best performance for both languages has been obtained by a 
combined learning system that used Maximum Entropy Models,
transformation-based learning, Hidden Markov Models as well as
robust risk minimization \cite{florian2003}.
Apart from the training data, this system also employed gazetteers and
the output of two externally trained named entity recognizers.
The performance of the system of Chieu et al. \shortcite{chieu2003}
was not significantly different from the best performance for English
and the method of Klein et al. \shortcite{klein2003} and the approach
of Zhang and Johnson \shortcite{zhang2003} were not significantly
worse than the best result for German.

Eleven teams have incorporated information other than the training
data in their system.
Four of them have obtained error reductions of 15\% or more for
English and one has managed this for German.
The resources used by these systems, gazetteers and externally trained
named entity systems, still require a lot of manual work.
Systems that employed unannotated data, obtained performance gains
around 5\%.
The search for an excellent method for taking advantage of the vast
amount of available raw text, remains open.

\section*{Acknowledgements}

Tjong Kim Sang is financed by IWT STWW as a researcher in the ATraNoS
project.
De Meulder is supported by a BOF grant supplied by the University of
Antwerp.

\begin{table}[t]
\begin{center}
\begin{tabular}{|l|c|c|c|}\cline{2-4}
\multicolumn{1}{l|}{English test}
                     & Precision & Recall  & F$_{\beta=1}$ \\\hline
Florian              & 88.99\%   & 88.54\% & 88.76$\pm$0.7 \\
Chieu                & 88.12\%   & 88.51\% & 88.31$\pm$0.7 \\
Klein                & 85.93\%   & 86.21\% & 86.07$\pm$0.8 \\
Zhang                & 86.13\%   & 84.88\% & 85.50$\pm$0.9 \\
Carreras (b)         & 84.05\%   & 85.96\% & 85.00$\pm$0.8 \\
Curran               & 84.29\%   & 85.50\% & 84.89$\pm$0.9 \\
Mayfield             & 84.45\%   & 84.90\% & 84.67$\pm$1.0 \\
Carreras (a)         & 85.81\%   & 82.84\% & 84.30$\pm$0.9 \\
McCallum             & 84.52\%   & 83.55\% & 84.04$\pm$0.9 \\
Bender               & 84.68\%   & 83.18\% & 83.92$\pm$1.0 \\ 
Munro                & 80.87\%   & 84.21\% & 82.50$\pm$1.0 \\
Wu                   & 82.02\%   & 81.39\% & 81.70$\pm$0.9 \\
Whitelaw             & 81.60\%   & 78.05\% & 79.78$\pm$1.0 \\
Hendrickx            & 76.33\%   & 80.17\% & 78.20$\pm$1.0 \\
De Meulder           & 75.84\%   & 78.13\% & 76.97$\pm$1.2 \\
Hammerton            & 69.09\%   & 53.26\% & 60.15$\pm$1.3 \\\hline
Baseline             & 71.91\%   & 50.90\% & 59.61$\pm$1.2 \\\hline
\multicolumn{4}{c}{}\\\cline{2-4}
\multicolumn{1}{l|}{German test}
                     & Precision & Recall  & F$_{\beta=1}$ \\\hline
Florian              & 83.87\%   & 63.71\% & 72.41$\pm$1.3 \\
Klein                & 80.38\%   & 65.04\% & 71.90$\pm$1.2 \\
Zhang                & 82.00\%   & 63.03\% & 71.27$\pm$1.5 \\
Mayfield             & 75.97\%   & 64.82\% & 69.96$\pm$1.4 \\
Carreras (b)         & 75.47\%   & 63.82\% & 69.15$\pm$1.3 \\
Bender               & 74.82\%   & 63.82\% & 68.88$\pm$1.3 \\
Curran               & 75.61\%   & 62.46\% & 68.41$\pm$1.4 \\
McCallum             & 75.97\%   & 61.72\% & 68.11$\pm$1.4 \\
Munro                & 69.37\%   & 66.21\% & 67.75$\pm$1.4 \\
Carreras (a)         & 77.83\%   & 58.02\% & 66.48$\pm$1.5 \\
Wu                   & 75.20\%   & 59.35\% & 66.34$\pm$1.3 \\
Chieu                & 76.83\%   & 57.34\% & 65.67$\pm$1.4 \\
Hendrickx            & 71.15\%   & 56.55\% & 63.02$\pm$1.4 \\
De Meulder           & 63.93\%   & 51.86\% & 57.27$\pm$1.6 \\
Whitelaw             & 71.05\%   & 44.11\% & 54.43$\pm$1.4 \\
Hammerton            & 63.49\%   & 38.25\% & 47.74$\pm$1.5 \\\hline
Baseline             & 31.86\%   & 28.89\% & 30.30$\pm$1.3 \\\hline
\end{tabular}
\end{center}
\caption{
Overall precision, recall and F$_{\beta=1}$ rates obtained by the
sixteen participating systems on the test data sets for the two
languages in the CoNLL-2003 shared task.
}
\label{tab-results}
\end{table}

\end{document}